\pdfoutput=1
\pdfoutput=1
\documentclass[11pt, a4paper]{article}

\usepackage[utf8]{inputenc}
\usepackage[T1]{fontenc}
\usepackage{textcomp}
\usepackage{graphicx}
\usepackage[english]{babel} 
\usepackage{times}
\usepackage{fancyhdr}
\usepackage{siunitx}
\DeclareSIUnit{\USD}{\$}

\usepackage{booktabs, tabularx, ragged2e}
\newcolumntype{C}{>{\centering\arraybackslash}X}
\newcolumntype{Y}{>{\RaggedRight\arraybackslash}X}

\usepackage[super]{cite}
\usepackage{microtype} 
\usepackage{xurl} 
\usepackage[colorlinks=true, allcolors=blue]{hyperref} 
\usepackage{geometry}
\usepackage{authblk}

\title{\textbf{Berta: an open-source, modular tool for AI-enabled clinical documentation}}
\author[1]{Samridhi Vaid}
\author[2,3]{Mike Weldon}
\author[3]{Jesse Dunn}
\author[1]{Sacha Davis}
\author[3]{Kevin Lonergan}
\author[2,3]{Henry Li}
\author[2,3]{Jeffrey Franc}
\author[1,4,5]{Mohamed Abdalla}
\author[1]{Daniel C. Baumgart}
\author[2,3]{Jake Hayward}
\author[1,4,5,*]{J Ross Mitchell}
\affil[1]{Department of Medicine, University of Alberta, Edmonton, Alberta, Canada}
\affil[2]{Department of Emergency Medicine, University of Alberta, Edmonton, Alberta, Canada}
\affil[3]{Alberta Health Services, Edmonton, Alberta, Canada}
\affil[4]{Department of Computer Science, University of Alberta, Edmonton, Alberta, Canada}
\affil[5]{Alberta Machine Intelligence Institute, Edmonton, Alberta, Canada}
\affil[*]{Corresponding author: jmitche2@ualberta.ca}
\date{}

\begin{document}

\maketitle
\thispagestyle{empty}

\begin{abstract}

\textbf{Background} 
Commercial AI scribes cost \$99-600 per physician per month, operate as opaque systems, and do not return data to institutional infrastructure, limiting organizational control over data governance, quality improvement, and clinical workflows.

\textbf{Methods} 
We developed Berta, an open-source modular scribe platform for AI-enabled clinical documentation, and deployed a customized implementation within Alberta Health Services (AHS) integrated with their existing Snowflake AI Data Cloud infrastructure. The system combines automatic speech recognition with large language models while retaining all clinical data within the secure AHS environment.

\textbf{Results} 
During eight months (November 2024 to July 2025), 198 emergency physicians used the system in 105 urban and rural facilities, generating \num{22148} clinical sessions and more than \num{2800} hours of audio. The use grew from 680 to \num{5530} monthly sessions. Operating costs averaged less than \$30 per physician per month, a 70–95\% reduction compared to commercial alternatives. AHS has since approved expansion to 850 physicians.

\textbf{Conclusions} 
This is the first provincial-scale deployment of an AI scribe integrated with existing health system infrastructure. By releasing Berta as open source, we provide a reproducible, cost-effective alternative that health systems can adapt to their own secure environments, supporting data sovereignty and informed evaluation of AI documentation technology.

\end{abstract}

\section*{Introduction}
The administrative burden of clinical documentation has become a defining challenge in modern healthcare. Physicians now spend nearly two hours on electronic health records for every hour of direct patient care~\cite{sinsky2016allocation}, with documentation that frequently extends into personal hours\cite{robertson2017electronic, sinsky2016allocation, arndt2017tethered}---a phenomenon dubbed "pajama time"~\cite{saag2019pajama}. This burden is not merely an inconvenience: excessive documentation demands are consistently associated with physician burnout~\cite{shanafelt2016relationship}, reduced job satisfaction~\cite{shanafelt2016relationship}, and workforce attrition~\cite{melnick2021analysis, han2019burnout}. The consequences extend to patients, as time spent on documentation is time diverted from direct clinical care, communication, and shared decision-making~\cite{hill2013clicks}. These demands are particularly burdensome in high-acuity settings such as emergency departments~\cite{cornish2021covid, sungbun2023impact, lavoietremblay2022influence}, where time-sensitive clinical decisions must compete with the requirement to produce prompt, thorough records.

Ambient AI scribes---systems that passively capture clinical conversations, transcribe speech, and generate structured documentation---have emerged as a promising response to this burden~\cite{watchlist2025}. Recent iterations of these tools combine automatic speech recognition with large language models to produce draft clinical notes that physicians can review and edit, reducing the cognitive and temporal costs of documentation. Pilot scribe projects have become mainstream in recent years, with products from vendors such as Ambience Healthcare, Doximity, and Abridge now deployed in routine clinical practice\cite{dai2025policy}. Early evidence suggests these tools can meaningfully reduce documentation time~\cite{hess2015scribe}, increase patient throughput~\cite{walker2019impact}, improve patient satisfaction and team communication~\cite{bastani2014ed, shuaib2019impact}, generating substantial institutional interest and investment.


However, the commercial landscape presents significant barriers to equitable adoption~\cite{sasseville2025impact, poon2025adoption, hassan2024barriers}. Subscription costs of \$99 to over \$600 per physician per month~\cite{scribeberry_cost, heidi_blog_cost} place these tools beyond reach for resource-constrained health systems, rural settings, and low- and middle-income countries. Commercial AI scribes operate as opaque systems~\cite{kim2025transparency} where clinical data is processed on vendor-controlled infrastructure, limiting institutional control over data governance, privacy compliance, and secondary use for quality improvement or research~\cite{england2025guidance, wong2025bridging}. Organizations cannot audit model behavior, customize outputs to local workflows, or retain generated data assets. This opacity also prevents healthcare providers from understanding the underlying technology, limiting their ability to make informed decisions about adoption or to develop institutional expertise. This dependency introduces strategic risks including pricing escalations, service discontinuation, and vendor lock-in as switching costs accumulate.


We developed Berta, an open-source, modular platform for AI-enabled clinical documentation. The system deploys entirely within institutional infrastructure, preserving data sovereignty while allowing organizations to customize and audit their own speech recognition and language model components. In partnership with Alberta Health Services (Canada's largest provincially integrated health system at implementation), we deployed Berta integrated with existing enterprise infrastructure. Here, we describe the system architecture, report on eight months of operational use by emergency physicians across 105 urban and rural facilities, and release the codebase to provide a reproducible, cost-effective foundation for health systems deploying AI documentation technology in secure environments.

\section*{Methods}
\subsection*{Co-Design, Development, and Architectural Philosophy}
The AI scribe was co-designed by a team of four emergency physicians, two machine learning engineers, one project analyst, and two supervising professors using an iterative, user-centred development process. The clinical team used the Alberta Health Services (AHS) deployment in routine practice and provided structured feedback during weekly meetings, which supported ongoing updates to match everyday emergency department workflows. Refinements validated during the AHS pilot were incorporated into the open-source Berta release.

\subsection*{Technical Architecture}
\begin{figure}[p]
    \centering
    \includegraphics[width=0.9\textwidth]{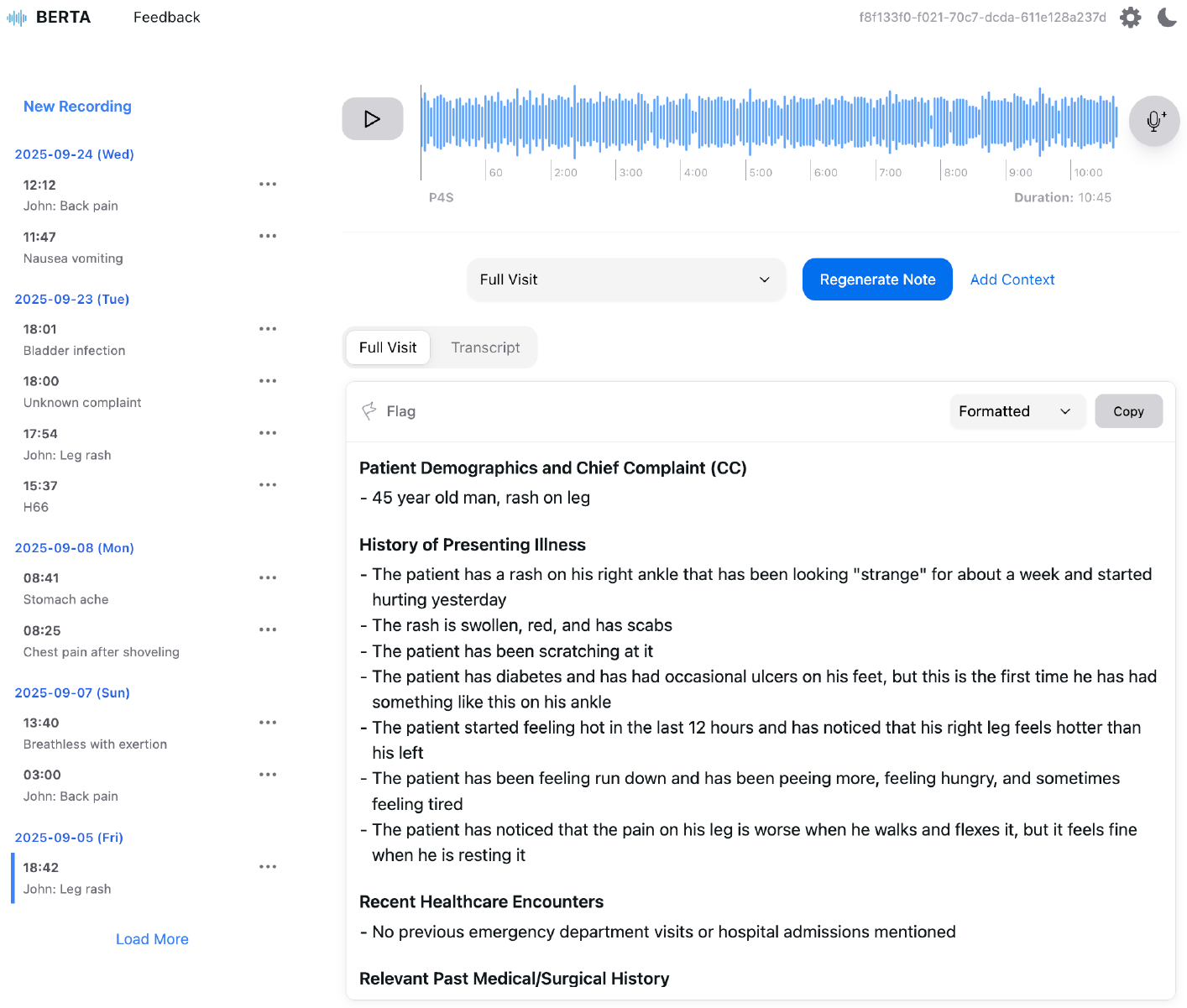}
    \caption{The Berta user interface. The left sidebar shows a chronological list of sessions, while the main area displays the audio waveform and the automatically generated medical note below. All data shown are from simulated patient sessions to protect privacy; no real patient information is used.}
    \label{fig:user_interface}
\end{figure}

\begin{figure}[p]
    \centering
    \includegraphics[width=0.9\textwidth]{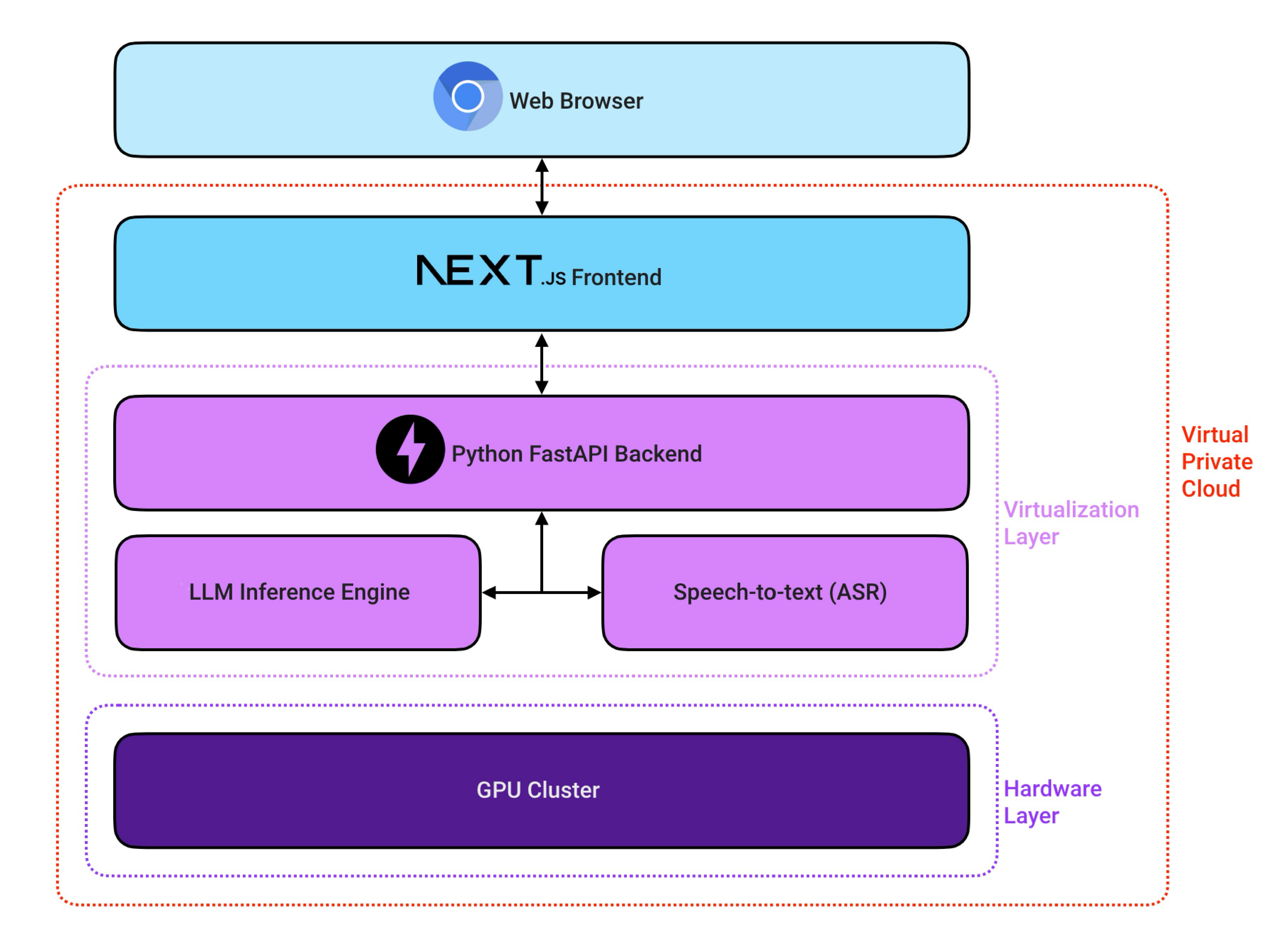} 
    \caption{Berta System Architecture. The system features a multi-layer design with a Next.js front-end and a Python FastAPI backend coupled with an LLM inference engine and a speech-to-text module. Multiple inference engines are supported including: vLLM, Ollama, LM Studio, and any OpenAI compatible API. Multiple speech-to-text modules are also supported including: WhisperX, Amazon Transcribe, Nvidia Parakeet V2, and others. Berta supports on-premises or cloud-based GPU acceleration and can be deployed securely within a virtual private cloud, with support for multiple user authentication systems. Icons courtesy of Wikicommons.}
    \label{fig:architecture}
\end{figure}

Berta comprises a Next.js (Vercel Inc., San Francisco, CA, USA) front-end (Figure~\ref{fig:user_interface}) and a FastAPI\cite{ramirez_fastapi} backend that exposes RESTful APIs for application logic, data processing, and integration (Figure~\ref{fig:architecture}).

In routine use, clinicians create a session in the web or mobile application and record or upload audio from a patient encounter. The system transcribes speech with an automatic speech recognition (ASR) model and then uses a large language model (LLM) to generate a structured draft clinical note from the transcript using configurable note templates (e.g., full visit, narrative, handover); users can also create and save custom templates. Clinicians review and edit the generated note before transferring it to the electronic health record (EHR).

The platform is modular across ASR and LLM components. Supported ASR backends include WhisperX~\cite{bain2023whisperx}, OpenAI Whisper~\cite{radford2023robust}, NVIDIA Parakeet via MLX~\cite{parakeet_mlx_impl, mlx2023, parakeet_tdt_0_6b_v2_nvidia}, and Amazon Transcribe~\cite{aws_transcribe_site}; supported LLM backends include local engines (Ollama~\cite{ollama_github}, vLLM~\cite{kwon2023efficient}, LM Studio~\cite{lm_studio_site}) and commercial endpoints (OpenAI API~\cite{openai_api_platform}, Amazon Bedrock~\cite{aws_bedrock_site}). Deployments can run from a single workstation to a GPU server within a secure virtual private cloud or enterprise cloud environments (e.g., Snowflake AI Data Cloud; Amazon Web Services), enabling institution-controlled data governance.

\subsection*{Pilot Implementation}
In November 2024, Berta was used as the template for a closed-source AI scribe deployed at Alberta Health Services (AHS), Canada's largest integrated health system; internally, it is known as Jenkins'' or the AHS Digital Scribe''. The protocol was approved by the Health Research Ethics Board (Health Panel), University of Alberta (Study ID Pro00138648; approval 11 July 2024). The pilot ran on on-premises Snowflake infrastructure and used WhisperX~\cite{bain2023whisperx} (Whisper extension~\cite{radford2023robust}) for transcription and a Snowflake-provisioned private-cloud GPT-4o~\cite{hurst2024gpt} for note generation.
All clinical audio and text remained within the AHS firewall and were not exported. Usage metrics reported here were derived from these operational data; the interface screenshot (Figure~\ref{fig:user_interface}) uses simulated patient data.
\subsection*{Pilot Data Collection and Analysis}
During the pilot, the system automatically recorded usage and session metadata. A ``session'' was defined as the set of one or more recordings, transcripts, and generated notes associated with a single patient encounter; multiple recordings and notes could be created within the same session. Physician shift data were obtained from the AHS electronic health record system.
\subsection*{Role of the Funding Source}
Sponsors had no role in study design, data collection, data analysis, data interpretation, writing of the report, or the decision to submit for publication.


\section*{Results}
Between November 2024 and July 2025, the system was used for \num{22148} sessions by 198 emergency physicians across 105 urban and rural facilities. Monthly volume increased from \num{680} sessions (November 2024) to \num{5530} (July 2025). Mean recorded session length was \num{7.6} minutes, yielding more than \num{2800} hours of clinical audio; 42\% of users customized at least one documentation template.

Based on Snowflake service consumption over the same period, operating costs averaged less than \SI{30}{\USD} per physician per month. This estimate was derived from web application and transcription container servers, LLM tokens, and storage. Exact per-user costs are difficult to calculate because the system runs on shared enterprise Snowflake infrastructure rather than a dedicated environment, with some costs directly related to consumption rather than number of users. Although costs will vary with scale and model selection, the pilot demonstrates that high-volume clinical use can be sustained at relatively low per-physician cost. Based on pilot outcomes, AHS approved expansion to 850 physicians.

\section*{Discussion}
This work demonstrates that AI-enabled clinical documentation can be deployed on a provincial scale within existing health system infrastructure, without relying on commercial AI scribe vendors. In partnership with Alberta Health Services, we developed and deployed an open-source AI scribe that has been used by 198 emergency physicians in 105 urban and rural facilities over an eight-month period. To our knowledge, this represents the first deployment of its kind integrated with a health system's existing data infrastructure. By releasing the underlying code base, we provide a reproducible foundation for health systems seeking to evaluate or implement AI documentation tools within their own secure environments.

\subsection*{Advantages of This Approach}
The integration of Berta with AHS's existing Snowflake infrastructure leads to several advantages over commercial alternatives. Most importantly, all clinical audio and text data remain within the institutional firewall, ensuring compliance with jurisdictional privacy requirements. Retention of data within institutional infrastructure also creates opportunities that are unavailable when data reside on vendor servers. Furthermore, \num{2800} hours of clinical audio collected during this deployment constitute a research asset for quality improvement, including analysis of communication patterns and systematic evaluation of transcription accuracy. This feedback loop enables continuous, locally driven improvement rather than dependence on vendor priorities. 


Local control extends to system customization and oversight; templates, prompts, and terminology specific to regional facilities and clinical workflows (facility names, physician names, regional slang, Indigenous community names) can be modified or added directly by the organization, and audit trails remain fully within institutional control, enabling end-to-end traceability in the event of clinical or legal review. The financial implications for users are also substantial: operating costs using cloud computing during the pilot averaged less than \$30 per physician per month; a reduction of 70–95\%  compared to commercial solutions typically priced at \$99–600 per month\cite{heidi_blog_cost}. A technically inclined physician could clone the repository and run a personal Berta-based scribe for \$0/month, using locally available compute and a quantized model. Because the system operates under an open-source license, there is no exposure to vendor pricing changes, service discontinuation, or contract re-negotiations.

The successful integration of AI scribe technology is based on both technical adaptability and well-defined assurance processes. Berta's design features, including on-premises processing, transparent data flows, human-in-the-loop review, and auditable logs, are consistent with NHS England's 2025 guide on ambient clinical documentation~\cite{england2025guidance}, which addresses risk assessment, regulatory classification, data security, and interoperability standards. Although we did not conduct formal NHS-specific evaluations, Berta's architecture supports these requirements in regions with similar regulatory frameworks.

\subsection*{Implications}
These findings have implications at many levels of healthcare. For health systems most broadly, this deployment establishes that implementation of an AI scribe in-house at scale is technically and financially feasible. For those looking to deploy their own Scribe, the open-source code base provides a working starting point that would substantially reduce development efforts. 
For clinical informatics and healthcare research groups, Berta offers a modular platform to study physician-AI interaction, documentation quality, and clinical communication, with access to underlying data that commercial deployments typically do not allow. For electronic health record vendors, the open-source architecture presents an integration opportunity that does not require licensing proprietary technology. For policymakers and regulators concerned with data sovereignty, this work demonstrates that AI-enabled clinical tools can operate entirely within institutional boundaries. Finally, in resource-limited settings, the cost reduction lowers a significant barrier to adoption and provides a pathway to evaluate AI documentation technology without committing to expensive licensing arrangements.

\subsection*{Limitations}
Several limitations should be noted in the current implementation. At this time, Berta does not integrate directly with electronic health records; physicians must copy and paste generated notes into the EHR, introducing a manual step that commercial products with direct integration may avoid. The absence of commercial support means that organizations must allocate internal IT resources for deployment, maintenance, and troubleshooting. The development velocity depends on community involvement and institutional contribution rather than dedicated vendor teams, which could result in slower iteration, and open-source projects carry some fragmentation risk if governance structures are not maintained. Long-term sustainability depends on continued institutional and community investment rather than market-driven incentives: a consideration that organizations should weigh when planning adoption.

\subsection*{Future Directions}
AHS has approved an expansion of the system to 850 emergency physicians, with a sequential rollout designed to allow continued refinement based on clinician feedback. Formal evaluation of this expanded deployment will assess note creation time, note quality, system reliability, and patients seen per shift, comparing AI-generated notes against established documentation standards. Longer-term directions include the integration of transcripts with electronic health records for longitudinal retrieval and the exploration of richer contextual inputs (such as prior patient history) to improve note generation, although such capabilities will require careful attention to safety and privacy considerations. The findings of the expanded AHS deployment will inform ongoing updates to the open-source code base, ensuring that the broader community benefits from real-world evidence as it accumulates.

\section*{Conclusion}
AI-driven documentation tools are rapidly entering clinical practice, yet most available solutions operate as proprietary systems that limit organizational control over data, customization, and cost. This work provides evidence that an alternative model is viable: an open-source AI scribe, deployed at scale, integrated with existing health system infrastructure, at a fraction of the commercial cost. By releasing Berta publicly, we offer health systems a reproducible path to evaluate and deploy AI documentation technology on their own terms.

\clearpage 

\section*{Contributors}
\textbf{Conceptualization:} M.W., J.R.M. \textbf{Methodology:} S.V., J.D., K.L., J.R.M. \textbf{Software:} S.V., J.D., K.L., J.R.M \textbf{Validation:} M.W., J.H., H.L., J.F. \textbf{Formal Analysis:} S.V. \textbf{Investigation:} M.W., J.H., H.L., J.F. \textbf{Resources:} AHS. \textbf{Data Curation:} S.V. \textbf{Writing – Original Draft:} S.V., J.R.M. \textbf{Writing – Review \& Editing:} All authors. \textbf{Supervision:} J.R.M., D.C.B. \textbf{Project Administration:} K.L., J.H., M.W., J.R.M. \textbf{Funding Acquisition:} J.H., M.W., J.R.M.

\section*{Data and Code Availability}
The Berta AI scribe source code is available under the Apache 2.0 license at \url{https://github.com/phairlab/berta-ai-scribe}. Due to patient privacy and institutional policies, individual-level clinical data from the Alberta Health Services deployment are not publicly available; only aggregate, de-identified metrics are reported in this article.

\section*{Medical disclaimer}
The Berta system is provided as a support tool only and is not intended to substitute for professional medical judgment. Healthcare providers maintain full responsibility for all clinical decisions and documentation.

\section*{Acknowledgments}
This work was supported by the Canadian Medical Association, MD Financial Management, and Scotiabank through the Health Care Unburdened Grant program. 
We acknowledge the support provided by the Canadian Institute for Advanced Research, the University Hospital Foundation, Alberta Health Services, Amazon Web Services, and Denvr Dataworks (\url{https://www.denvrdata.com}). 
This project uses third-party libraries and models, including WhisperX (BSD 2-Clause) and Meta Llama 3 (Meta Platforms, Inc., Menlo Park, CA, USA; Meta Llama 3 Community License). This project is ``Built with Meta Llama 3'' and complies with the Acceptable Use Policy of Meta Llama 3. We also thank the open-source developer communities behind WhisperX, NVIDIA Parakeet (CC-BY-4.0), vLLM (Apache 2.0), Ollama (MIT License), LM Studio, and related projects for making their tools available.

\bibliographystyle{unsrt}
\bibliography{sample}

\clearpage

\sisetup{separate-uncertainty, table-number-alignment=center}

\end{document}